%% file: main.tex
\documentclass[runningheads]{llncs}
\usepackage{graphicx}
\usepackage{todonotes}
\usepackage{nicematrix}
\usepackage{pifont}
\usepackage{mathrsfs}
\usepackage{url}

\input{macro}

\pgfplotsset{compat=1.17} 
\begin{document}

\title{Guiding Enumerative Program Synthesis with Large Language Models}

%
%
\author{Yixuan Li\inst{1}\orcidID{0009-0007-4619-3476} \and
Julian Parsert\inst{1}\inst{2}\inst{3}\orcidID{0000-0002-5113-0767} \and
Elizabeth Polgreen\inst{1}\orcidID{0000-0001-9032-7661}}
%
%
\institute{University of Edinburgh, UK\\
\email{\{yixuan.li.cs,~elizabeth.polgreen\}@ed.ac.uk}\\
\and
University of Oxford, UK and University of Innsbruck, Austria\\ 
\email{julian.parsert@gmail.com}
}

\maketitle              

\begin{abstract}
Pre-trained Large Language Models (LLMs) are beginning to dominate the discourse around automatic code generation with natural language specifications. In contrast, the best-performing synthesizers in the domain of formal synthesis with precise logical specifications are still based on enumerative algorithms. 
In this paper, we evaluate the abilities of LLMs to solve formal synthesis benchmarks by carefully crafting a library of prompts for the domain. When one-shot synthesis fails, we propose a novel enumerative synthesis algorithm, which integrates calls to an LLM into a weighted probabilistic search. This allows the synthesizer to provide the LLM with information about the progress of the enumerator, and the LLM to provide the enumerator with syntactic guidance in an iterative loop.
We evaluate our techniques on benchmarks from the Syntax-Guided Synthesis (SyGuS) competition. We find that GPT-3.5 as a stand-alone tool for formal synthesis is easily outperformed by state-of-the-art formal synthesis algorithms, but our approach integrating the LLM into an enumerative synthesis algorithm shows significant performance gains over both the LLM and the enumerative synthesizer alone and the winning SyGuS competition tool.
\end{abstract}

\input{sections/01-intro}

\input{sections/03-background}
\input{sections/04-overview}

\input{sections/05-method1}
\input{sections/06-method2}

\input{sections/07-eval}

\input{sections/08-threats}

\input{sections/02-related}
\input{sections/09-conclusion}

\paragraph{Acknowledgements:} This work was in part supported by an Amazon Research Award, a Royal Academy of Engineering research fellowship, and the European Research Council (ERC)
project FormalWeb3 (Grant ID 101156734).

%
%
\bibliographystyle{splncs04}
\bibliography{LLM-SyGuS}


\end{document}

%% file: macro.tex
\usepackage{tikz}
\usepackage{listings}
\usepackage{algorithm}
\usepackage[noend]{algpseudocode}
\usepackage{amsmath}
\usepackage{amssymb}
\usepackage{enumitem}
\usepackage{prettyref}
\usepackage{multirow}
\usepackage{diagbox}
\usepackage{textcomp}
\usepackage{inconsolata}
\usepackage{wrapfig}
\usepackage{xspace}
\usepackage{regexpatch}
\usepackage{listings}
\newrefformat{sec}{Section~\ref{#1}}
\usepackage{xcolor}
\usepackage{pgfplots}
\usepackage{float}
\usepackage{tikz}
\tikzset{
  Box/.style={
    rectangle, 
    rounded corners, 
    minimum width=2.5cm, 
    minimum height=1.25cm,
    text centered, 
    draw=black
    },
  arrow/.style={
    thick,
    ->,
    >=stealth
  }
}

\definecolor{bar1}{HTML}{8741a6}
\definecolor{bar2}{HTML}{e36b3a}
\definecolor{bar3}{HTML}{f5af3b}

\newcommand{\methodone}{\pretrainedPCFG}
\newcommand{\methodtwo}{\integratedLLM}
\newcommand{\pretrainedPCFG}{pCFG-synth\xspace}
\newcommand{\integratedLLM}{iLLM-synth\xspace}
\newcommand{\prob}{\mathbb{P}}
\newcommand{\weight}{w}
\newcommand{\tree}{\mathcal{T}}
\newcommand{\edgeWeight}{\omega}

\newcommand{\languageOfGrammar}[1]{\ensuremath{\mathcal{L}^#1}}

\makeatletter
\algdef{SE}[REPEATN]{RepeatN}{EndRepeatN}[1]{\algorithmicrepeat\ #1 \textbf{times}}{\algorithmicend \textbf{ repeat}}

\DeclareMathOperator*{\kleeneStar}{\!^\ast}

\newcounter{prompt}
\newcounter{response}
\newcounter{program}
\newcounter{benchmark}

\lst@UserCommand\lstlistofprompt{\bgroup
    \let\contentsname\lstlistprompt
    \let\lst@temp\@starttoc \def\@starttoc##1{\lst@temp{lor}}%
    \tableofcontents \egroup}
\lstnewenvironment{prompt}[1][]{%
  \renewcommand{\lstlistingname}{Prompt}%
  \let\c@lstlisting=\c@prompt
  
  \xpatchcmd*{\lst@MakeCaption}{lol}{lor}{}{}%
\lstset{language=text,
  captionpos=b,
  #1}}
  {}

\lst@UserCommand\lstlistofresponse{\bgroup
    \let\contentsname\lstlistresponse
    \let\lst@temp\@starttoc \def\@starttoc##1{\lst@temp{lop}}%
    \tableofcontents \egroup}
\lstnewenvironment{response}[1][]{%
  \renewcommand{\lstlistingname}{Response}%
  \let\c@lstlisting=\c@response
    
  \xpatchcmd*{\lst@MakeCaption}{lol}{lop}{}{}%
\lstset{language=text,
  captionpos=b,
  #1}}
  {}

\lst@UserCommand\lstlistofprogram{\bgroup
    \let\contentsname\lstlistprogram
    \let\lst@temp\@starttoc \def\@starttoc##1{\lst@temp{lop}}%
    \tableofcontents \egroup}
\lstnewenvironment{program}[1][]{%
  \renewcommand{\lstlistingname}{Program}%
  \let\c@lstlisting=\c@program
    
  \xpatchcmd*{\lst@MakeCaption}{lol}{lop}{}{}%
\lstset{language=text,
  captionpos=b,
  #1}}
  {}

\lst@UserCommand\lstlistofbenchmark{\bgroup
    \let\contentsname\lstlistbenchmark
    \let\lst@temp\@starttoc \def\@starttoc##1{\lst@temp{lop}}%
    \tableofcontents \egroup}
\lstnewenvironment{benchmark}[1][]{%
  \renewcommand{\lstlistingname}{Example}%
  \let\c@lstlisting=\c@benchmark
    
  \xpatchcmd*{\lst@MakeCaption}{lol}{lop}{}{}%
\lstset{language=text,
  captionpos=b,
  #1}}
  {}

\lstdefinelanguage{text}{
    basicstyle=\ttfamily\scriptsize,
    sensitive = false,
    keywords={},
    numbers=none,
    numberstyle=\ttfamily\scriptsize,
    stepnumber=1,
    numbersep=8pt,
    showstringspaces=false,
    breaklines=true,  
    linewidth=0.95\linewidth,
    xleftmargin=0.05\linewidth
    }

%% file: sections/01-intro.tex
\section{Introduction}
Program synthesis is the task of automatically generating programs that satisfy a given specification. It has applications in planning~\cite{planning-sygus}, program analysis~\cite{program-synthesis-for-analysis}, data-wrangling~\cite{dillig-wrangling} and more. The dominant techniques for formal program synthesis are based around enumeration~\cite{cvc4sy,eusolver,dryadsynth}, and a key challenge is how to guide this enumeration to search a huge space of possible programs efficiently. Syntax-Guided Synthesis(SyGuS)~\cite{alur2013syntax} allows the user to restrict the space of possible programs using a context-free grammar, and, in later work, this has been extended using pre-trained probabilistic models such as higher-order grammars~\cite{euphony} and neural networks~\cite{grammarfiltering}, trained on a dataset of solved synthesis problems. However, obtaining these datasets for pre-training is challenging. 

In parallel, the use of pre-trained large language models (LLMs) to generate code is rapidly gaining traction, with impressive results being obtained on benchmarks with natural language specifications and input-output examples~\cite{chen2021evaluating}. These benchmarks are very different in style to the logical specifications that formal program synthesis tackles, as most are procedural code, in Python, and solve classic programming exercise questions that might be asked of students or interview candidates, and that one may find in abundance on sources used in training data such as StackOverflow and GitHub. In contrast, formal program synthesis benchmarks, such as those in the SyGuS competition, require functional code, which must satisfy precise logical specifications derived from problems such as program analysis~\cite{program-synthesis-for-analysis}, and are certainly less abundant in sources of publicly available code for training machine learning models.

In this paper, we set out to investigate whether off-the-shelf large language models can solve formal program synthesis problems. We craft a library of prompts, which enables us to solve roughly $50\%$ of the SyGuS competition benchmarks. We hypothesize that, in the cases where the LLM returns only incorrect solutions, the correct solutions are most often in the vicinity of the incorrect solutions, and that, by searching in the neighborhood of the incorrect solutions, we may be able to guide an enumerative synthesizer to find a solution faster. To that end, we construct a probabilistic Context-Free Grammar (pCFG) based on the incorrect solutions proposed by the LLM, and use this to guide an enumerative synthesizer within a CounterExample Guided Inductive Synthesis (CEGIS) loop. 

Our final contribution is a full integration of these techniques in a novel CEGIS algorithm with an inline syntactic oracle, in the form of an LLM that is queried by an enumerative synthesis phase. We incorporate information obtained during the synthesis search into the queries, prompting the LLM with partially enumerated functions, incorrect solutions, and counterexamples, and requesting that it provide ``helper functions'', which we use to update the pCFG guiding the enumerator. 

We implement all three techniques described above and evaluate them on benchmarks from the Syntax-Guided Synthesis competition. We compare with two baselines: the first is an enumerative synthesizer where all rules in the grammar are given equal likelihood, and the second is cvc5~\cite{cvc5}, the state-of-the-art SyGuS solver. All techniques easily outperform the baseline enumerator, and the final technique outperforms cvc5. Our results demonstrate that, whilst large language models do have the potential to make significant contributions in the domain of formal program synthesis, this can currently only be achieved by combining these techniques with existing algorithms in the literature. Enumerative synthesis is not yet obsolete!

The main contributions of our work are as follows:
A set of prompts for prompting a pre-trained Large Language Model to solve formal program synthesis problems (Section~\ref{sec:prompting});
A method for guiding an enumerative synthesizer using LLM-generated probabilistic context-free grammars (Section~\ref{sec:pCFG});
A novel approach to integrating an LLM into an enumerative synthesizer (Section~\ref{sec:illm}); And, finally,
an implementation and evaluation of all of the above on benchmark problems taken from the Syntax-Guided Synthesis competition. The results outperform cvc5, the state-of-the-art synthesizer, as well as our baseline enumerators.

%% file: sections/03-background.tex
\section{Background}
Program synthesis focuses on automated program creation that satisfies a high-level specification, which can be comprehensive, such as a basic, unrefined program, or incomplete, like a logical formula or a set of test cases.

\begin{definition}[Context-Free Grammar, CFG]\label{def:cfg}
A context-free grammar is a 4-tuple $G = (V, \Sigma, R, S)$. $V$ is a finite set of variables also known as non-terminal symbols. $\Sigma$ with $\Sigma \cap V = \emptyset$ is called the set of terminal symbols or alphabet. $R \subseteq V \times (V \cup \Sigma)\kleeneStar$ is a finite relation describing the production rules of the grammar.  We define $R_\Sigma = R \cap V\times \Sigma\kleeneStar$, i.e. the set of rules restricted to those whose right-hand side only consists of terminal symbols.  
Elements of $(V \cup \Sigma)\kleeneStar$ are known as words in sentential form.
$S \in V$ is the start symbol of the grammar $G$. 
\end{definition}
Given a context-free grammar $G = (V, \Sigma, R, S)$ with $x,y \in (V \cup \Sigma)\kleeneStar$ and $(\alpha,\beta) \in R$ we say that $x\alpha y$ yields $x\beta y$, written $x\alpha y \rightarrow x\beta y$. We say that $x$ derives $y$ written $x \rightarrow\kleeneStar y$ if either $x = y$ or $x \rightarrow x_1 \rightarrow \dots x_{n} \rightarrow y$ for $n\geq 0$.
Finally, we define the \emph{language} of a grammar $\languageOfGrammar{G} = \{s \in \Sigma\kleeneStar \mid S \rightarrow\kleeneStar s \}$. We now introduce two extensions of context-free grammars:
\begin{definition}[Weighted Context-Free Grammar, wCFG]\label{def:wcfg}
A weighted context-free grammar(wCFG)~\textsc{\cite{liang2010learning,menon2013machine}} is a 5-tuple $W_G = (V, \Sigma, R, S, W)$ such that $(V, \Sigma, R, S)$ is a context-free grammar and $W$ is a function assigning a numeric value to each rule $r \in R$. 
\end{definition}
\begin{definition}[Probabilistic Context-Free Grammar, pCFG]\label{def:pcfg}
A probabilistic context-free grammar~\textsc{\cite{liang2010learning,menon2013machine}} is a 5-tuple $P_G = (V, \Sigma, R, S, \prob)$ such that $(V, \Sigma, R, S)$ is a context-free grammar and $\prob$ is a probability mass function assigning a probability $\prob[r]$ to each rule $r \in R$. 
$\prob_\Sigma$ is the probability mass function that assigns a probability to $\prob_\Sigma[r]$ to each rule $r \in R_\Sigma$. 
A pCFG is a specific instance of a wCFG.
\end{definition}

In general, program synthesis is concerned with the generation (i.e., synthesis) of a program that satisfies a certain specification. Syntax-guided synthesis (SyGuS) describes a standardized function synthesis format that precisely defines a synthesis problem within first-order theories~\cite{DBLP:series/faia/BarrettSST21}. We will use the notation $\phi[F \mapsto f]$ to denote the replacing of all occurrences of $F$ in $\phi$ with $f$ while substituting
all arguments to $f$ by the arguments of $F$ in the same order. 
\begin{definition}[Syntax-Guided Synthesis, SyGuS]\label{def:sygus}
A SyGuS problem is a 4-tuple $\langle T,G,\phi,F\rangle$ such that $T$
is a first-order theory, $G$ is a context-free grammar, $\phi$ is a first-order
formula, and $F$ is a function symbol that may occur in $\phi$. A solution to a SyGuS problem
$\langle T,G,\phi,F\rangle$ is either a function $f$ such that
$T \models \phi[F \mapsto f]$ and $f \in \languageOfGrammar{G}$,
or proof that no such function can exist.
\end{definition}
SyGuS closely follows the syntax and semantics of SMT, and hence $T$ usually refers to
theories that are also common in SMT. Usually, SMT solvers are queried in the background of SyGuS solvers to verify solution candidates.
This connection is made explicit in \emph{Counter-Example Guided Inductive Synthesis} (CEGIS)~\cite{solar2006combinatorial}.
CEGIS is a family of algorithms that alternate between a synthesis phase, which searches for a candidate solution that works for a subset of inputs, and a verification phase, where the candidate is checked against all possible inputs. If the verification fails, a counterexample is passed back to the synthesis phase and appended to the subset of inputs used to guide the search. The synthesis phase is often implemented as an enumerative search.
An example SyGuS problem is shown in Example~\ref{benchmark:max3}.

\textbf{Generative Large Language Models} Generative Large Language Models (LLMs) are advanced Artificial Intelligence (AI) systems based on transformer models and trained on vast datasets to produce human-like text, followed by human-provided instruction prompts~\cite{brown2020language}. One application of LLMs is generating code from natural language specifications~\cite{chen2021evaluating}. 

%% file: sections/04-overview.tex
\section{Overview}
In this work, we first present a carefully tailored set of prompts that we use to evaluate an LLM's ability to solve formal synthesis problems. We construct an iterative loop where we prompt the LLM, verify the candidate solution, and if the solution fails, we prompt the LLM again.
\input{figures/methodone}

We then present two methods for integrating syntactic guidance from pre-trained LLMs into an enumerative CEGIS algorithm. The first method, shown in Figure~\ref{fig:method_1_diagram}, prompts an LLM for solutions to the benchmark, and generates a pCFG from these solutions before deploying an enumerative synthesizer, increasing the chance of the LLM solving the synthesis problem outright. We refer to this method as \pretrainedPCFG. The second method, shown in Figure~\ref{fig:method_2_diagram}, integrates the prompting within the enumerative synthesizer, allowing the prompts to incorporate additional information obtained during the synthesis process. Here, instead of asking the LLM to provide a full solution, we ask it to provide helper functions to help ``a student'' complete the partially enumerated program. We use the responses to augment the set of production rules in the grammar and update the weights across the existing production rules. We refer to this approach, which integrates an LLM into an enumerative synthesizer, as \integratedLLM. In this section, we give an overview of these two approaches. The details of the components of both approaches and their relative performances are found in the subsequent sections. We integrate both approaches with a probabilistic top-down enumerator and a weighted search based on the $A^*$ algorithm~\cite{astar,euphony}.

\input{prompts/max3}

\input{figures/methodtwo}

%% file: figures/methodone.tex
\begin{figure}[ht]
\centering
\begin{tikzpicture}[node distance=5.5cm] 
    \node (LLM) [Box] {\textbf{LLM}};
    \node (Verifier) [Box, right of=LLM] {\textbf{Verifier}};
    \node (Enumerator) [Box, below of=Verifier, yshift=3cm] {\textbf{Enumerator}};
    \node (output) [right of=Verifier, xshift=-2.5cm] {}; 
    \node (input) [left of=LLM, xshift=2cm] {}; 
    \node (input_enum) [left of=Enumerator, xshift=2cm] {}; 
    \draw [arrow] (Verifier) -- node[above]{Solution}(output);
    \draw [arrow] (input) -- node[above]{Prompt}(LLM);
    \draw [arrow] (input_enum) -- node[above]{Grammar}(Enumerator);
    \newcommand{\shft}{3mm} 
    
    \draw [arrow] ([yshift=\shft]LLM.east) -- ([yshift=\shft]Verifier.west) node[above,midway,text width=2cm]{~~Candidate\\~~~Program};
    \draw [arrow] ([yshift=-\shft]Verifier.west) -- ([yshift=-\shft]LLM.east) node[above,midway]{Prompts};
    \draw [arrow] ([xshift=\shft]Verifier.south) -- ([xshift=\shft]Enumerator.north) node[midway,midway,text width=4cm,xshift=.5cm,align=left]{~~~~~~~~~~~~~~~~~~~Rule\\~~~~~~~~~~~~~~~~~Weights};
    \draw [arrow] ([xshift=-\shft]Enumerator.north) -- ([xshift=-\shft]Verifier.south) node[left,midway,text width=4cm,xshift=-.1cm,align=right]{Candidate\\Program~~};
\end{tikzpicture}
\caption{An overview of \methodone.  Both the verifier and the LLM have access to the specification $\phi$ (which is used to generate the prompt for the LLM, as well as to check whether candidate programs are correct).}
\label{fig:method_1_diagram}
\end{figure}
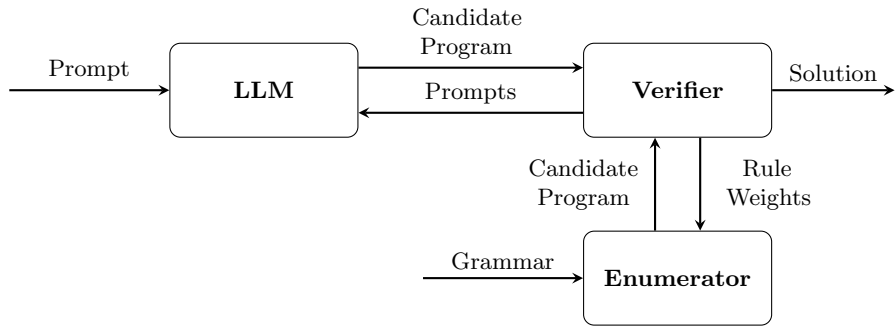

%% file: prompts/max3.tex
\begin{benchmark}[caption={A SyGuS specification that asks for a program that synthesizes the maximum of $3$ inputs. We omit some the grammar and variable declarations for brevity.},label={benchmark:max3}]
(set-logic LIA)
(synth-fun fn ((vr0 Int) (vr1 Int) (vr2 Int)) Int)
(constraint (>= (fn vr0 vr1 vr2) vr0))
(constraint (>= (fn vr0 vr1 vr2) vr1))
(constraint (>= (fn vr0 vr1 vr2) vr2))
(constraint (or (= vr0 (fn vr0 vr1 vr2)) (or (= vr1 (fn vr0 vr1 vr2)) (= vr2 (fn vr0 vr1 vr2)))))
(check-synth)
\end{benchmark}

%% file: figures/methodtwo.tex
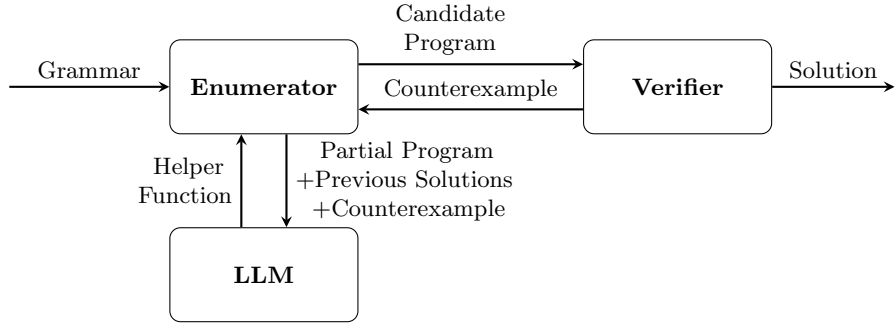
\begin{figure}
\centering
\begin{tikzpicture}[node distance=5.5cm] 
    \node (Enumerator) [Box] {\textbf{Enumerator}};
    \node (Verifier) [Box, right of=Enumerator] {\textbf{Verifier}};
    \node (LLM) [Box, below of=Enumerator, yshift=3cm] {\textbf{LLM}};
    \node (output) [right of=Verifier, xshift=-2.5cm] {}; 
    \node (input) [left of=Enumerator, xshift=2cm] {}; 
    \draw [arrow] (Verifier) -- node[above]{Solution}(output);
    \draw [arrow] (input) -- node[above]{Grammar}(Enumerator);
  
    \newcommand{\shft}{3mm} 
    \draw [arrow] ([yshift=\shft]Enumerator.east) -- ([yshift=\shft]Verifier.west) node[above,midway,text width=2cm]{Candidate\\~Program};
    \draw [arrow] ([yshift=-\shft]Verifier.west) -- ([yshift=-\shft]Enumerator.east) node[above,midway]{Counterexample};
    
    \draw [arrow] ([xshift=\shft]Enumerator.south) -- ([xshift=\shft]LLM.north) node[right,midway,text width=4cm,align=left]{~~~Partial Program\\+Previous~Solutions\\~~+Counterexample};
    \draw [arrow] ([xshift=-\shft]LLM.north) -- ([xshift=-\shft]Enumerator.south) node[left,midway,text width=2cm,align=right]{Helper~~\\Function};
\end{tikzpicture}
\caption{An overview of \methodtwo. Both the verifier and the enumerator have access to the specification $\phi$ (which is used to generate the prompt for the LLM, as well as to check whether candidate programs are correct)}
\label{fig:method_2_diagram}
\end{figure}

%% file: sections/05-method1.tex
\section{Stand-alone LLM}
\label{sec:standalone}

In this section, we describe how we prompt the LLM as a stand-alone synthesizer. These prompting techniques are then also deployed by \methodone. We use GPT-3.5-turbo as the LLM. Note that the model is not fine-tuned to this problem setting. Furthermore, we rename any functions and variables in the SyGuS benchmarks to generic names to avoid the LLM producing solutions solely based on the function names. 
\subsection{Prompting the LLM}
\label{sec:prompting}
We design a library of prompts for program synthesis problems with logical specifications and a single target function to synthesize. These prompts are deployed in an iterative loop, until a correct solution is obtained, or the library of prompts is exhausted. 

Prompting is an art rather than a science, but we hypothesize that it is better to ask the LLM to give a solution in a language that is more common in the training data, and then request it to translate it into our desired SMT-LIB, and experiment with both Python and Lisp. On a subset of 50 benchmarks, we observed that soliciting responses in Lisp resulted in a 6\% enhancement in the resolution of benchmarks compared to utilizing Python (and both were better than asking for SMT-lib directly). It is also reported in the literature that the efficacy of LLMs can be augmented by including emotional stimuli within prompts~\cite{emotional-prompts}. Incorporating the emotional prompt leads to an 8\% increase in the likelihood of generating accurate solutions compared to standard, non-emotional prompts in our study. Details of our initial prompting experiments are in the extended version of this paper~\footnote{\url{https://arxiv.org/html/2403.03997}}.

The following is an example prompt and response sequence for the LIA example shown in Example~\ref{benchmark:max3}:
\input{prompts/pcfg_prompt_1}
\input{prompts/pcfg_response_1}
\input{prompts/pcfg_prompt_2}

\subsubsection{Prompts for invariant synthesis}
Invariant synthesis is a specific instance of program synthesis: given a pre-condition $pre(x)$, transition-relation $trans(x, x')$ and post-condition $post(x)$, the synthesizer is required to provide an invariant $inv$ that satisfies the following constraint:
$
\forall x, x'. pre(x) \implies inv(x) \,\,\wedge 
    (inv(x) \wedge trans(x,x)) \implies inv(x') \,\,\wedge
    inv(x) \implies post(x).
$
    We find that LLMs struggle to reason about constraints presented in the above format. Inspired by ``chain-of-thought''~\cite{chain-of-thought} prompting, where the LLM is asked to provide a step-by-step explanation, we augment our prompting strategy for invariants by asking the LLM first to explain the constraints. 
    After requesting this explanation, we follow the same interactive prompt strategy as before.

\subsubsection{Lisp to SMT-LIB Converter}
The final prompts in our prompt library are to ask the LLM to convert any functions given in Lisp to correct SMT-LIB functions:
\input{prompts/convertor}
Upon receiving a response from the LLM, we extracted the Lisp program and subjected it to format verification. The resulting SMT-LIB code is represented:

\input{prompts/convertor_response}

\section{Synthesis with pCFG Guidance: \methodone}
\label{sec:updateweights}
We hypothesize that, if the LLM did not propose a correct solution, the correct solution is likely to be roughly in the same ``area'' as the incorrect solutions it suggested, and so our synthesis algorithm aims to prioritize this area when searching for candidate programs. For simplicity, we use a simple weighted Context-Free Grammar to represent the area of solutions proposed by the LLM. We then present methods for searching the space: the first is a probabilistic top-down search, shown in Algorithm~\ref{algo:pCFG-synth}; the second is based on an adaptation of the $A^*$ algorithm~\cite{astar,euphony}, and we integrate both into CEGIS searches as shown in Algorithm~\ref{algo:cegis}. The verification phase in Algorithm~\ref{algo:cegis} is implemented via a call to an SMT solver, which checks, for a candidate solution $f$, whether there exists an input such that the specification is violated, i.e., $\exists x. \neg\phi[F \mapsto f]$.
\input{algorithms/cegis}

\subsection{Inferring a Weighted CFG}
\label{sec:wcfg}
\label{sec:pCFG}
In this section, we describe how we infer a weighted Context-Free Grammar from the incorrect solutions produced by the large language model.

\begin{definition}[Derivations] Given a context-free grammar $G$, and a sentence $s$, the sentence is in the language of the grammar if $S \rightarrow\kleeneStar s$, where $S$ is the start symbol of the grammar. The derivation of $s$ from $S$ is a sequence of rules such that $S \xrightarrow{r_0} s_1 \xrightarrow{r_1}\ldots s_n \xrightarrow{r_n} s$ and $r_0 \ldots r_n \in R$. We denote the derivation of $s$ by the sequence of rules $r_0, \ldots r_n$ as $D_s = \{r_0, \ldots r_n\}$. The left-most derivation is a derivation such that all rules expand the left-most non-terminal symbol in the sentential form.
\end{definition}
 From here on in, all derivations are assumed to be the left-most derivation, and   
we assume the grammar is unambiguous, i.e., there exists a single left-most derivation for any sentence in the language.

Given a set of possible programs $prog \in \languageOfGrammar{G}$ generated by the language model, we calculate a weight for each rule $r_i \in R$ as the number of times that rule appears in the left-most derivations of the programs. That is, 
\begin{equation}
\weight[r_i]  = \sum_{prog_i \in prog} |r_i| \in D_{prog_i},
\label{eq:weightupdate}
\end{equation}
where $|r_i|$ is the number of times $r_i$ appears in the derivation.
For example, consider Response~\ref{prog:llm}: the weights are calculated as $\weight[r_1] = 3, \weight[r_2] = 3, \weight[r_3] = 3, \weight[r_4] =4, \weight[r_5] = 3$. These correspond to the rules from Example~\ref{benchmark:max3}:
\begin{align*}
    r_1&: \texttt{Start} \rightarrow \texttt{(ite StartBool Start Start)}\\
    r_2&: \texttt{Start} \rightarrow \texttt{vr0} \\
    r_3&: \texttt{Start} \rightarrow \texttt{vr1} \\
    r_4&: \texttt{Start} \rightarrow \texttt{vr2} \\
    r_5&: \texttt{StartBool} \rightarrow \texttt{(>= Start Start)}. \\
\end{align*}

\subsubsection{Probabilistic Context-Free Grammar}
Given a wCFG, we derive a simple pCFG by assuming that the probability associated with a rule $r_i \colon \alpha \rightarrow \beta$ is equal to the weight $\weight[\alpha \rightarrow \beta]$ of $r_i$, divided by 
$|\pi[\alpha]| = |\alpha \times (\Sigma \cup V)\kleeneStar \in R|$, i.e., the total number of rules that could be applied to $\alpha$. 
That is
$
\prob[\alpha \rightarrow \beta] = \frac{\weight[\alpha \rightarrow \beta]}{|\pi[\alpha]|}$.
By extension, $
\prob_\Sigma[\alpha \rightarrow \beta] = \frac{\weight[\alpha \rightarrow \beta]}{|\pi[\alpha]}|$ iff $\beta \in \Sigma$ and $0$ otherwise.

\subsection{Probabilistic Guided Search}

The aim of our algorithm is thus to search the area of programs closest to those with the highest weights in the wCFG, or highest probabilities in the corresponding pCFG. We adapt and implement two search methods for doing this: the first is a probabilistic top-down search. To this end, we first introduce the notion of a grammar tree.
\begin{definition}[Grammar tree]\label{def:grammar tree}
We represent the search space as a grammar tree. Given a context-free grammar $G = (V, \Sigma, R, S)$, the graph of sentential forms, or grammar tree, $\tree(G)$ defined inductively: $S$ is the root of the tree, and for all $x,y \in (V \cup \Sigma)\kleeneStar$ with $x \rightarrow y$ and $x$ being a node of the tree, then $y$ is a child node of $x$.
\end{definition}
To implement our probabilistic guided search, we extend this definition to a probabilistic grammar tree. Given a pCFG, $P_G = (V, \Sigma, R, S, \prob)$, a probabilistic grammar tree $\tree(P_G)$ is a directed labelled graph as defined before, but each edge has a corresponding weight $\edgeWeight$ given by $\prob$.
We limit the edges to only those needed for the left-most derivations, and so $\mathcal{E}$ and $\edgeWeight$ are defined as follows:
\begin{align*}
    &\mathcal{E} = \{x\alpha y \xrightarrow{\alpha \rightarrow \beta}x\beta y \,|\, \alpha \rightarrow \beta \in R, x \in \Sigma\kleeneStar, \alpha \in V, \beta, y \in (V  \cup \Sigma)\kleeneStar\},\\
    &\edgeWeight[\alpha \rightarrow \beta] = \prob[\alpha \rightarrow \beta].
\end{align*}
Note that this guarantees that, for any node, the sum of the weight on the edges leaving that node is equal to $1$.

\input{algorithms/enumerate}

We search this grammar tree using a top-down enumerative synthesizer, shown in Algorithm~\ref{algo:enumerate}. This enumerates possible programs in the grammar in a top-down manner, expanding non-terminals by randomly sampling from the categorical distribution over the production rules. That is, the search algorithm starts by considering the node corresponding to the start symbol $S$. It then chooses the next node by sampling from a categorical distribution with event probabilities corresponding to the probabilities on the outgoing edges of the current node. The categorical distribution is a generalization of the Bernoulli distribution and describes the possible results of a random variable that can take one of $K$ possible categories, with the probability of each category separately specified. Formally, to sample a rule $\alpha \times \beta$ to apply to a non-terminal symbol $\alpha$, we sample from the distribution:
$$
(\alpha \times \beta) \sim Cat(|\pi[\alpha]|,\{\prob[\pi[\alpha]_1], \prob[\pi[\alpha]_2],\ldots\}),
$$
where $|\pi[\alpha]|$ is the number of rules that could be applied to $\alpha$ and $\pi[\alpha]_i$ is the $i^{th}$ of those rules, and $\{\prob[\pi[\alpha]_1], \prob[\pi[\alpha]_2],\ldots\}$ is a vector of probabilities corresponding to those rules.

We then apply the sampled rule, and repeat the process. We use $prog.\{\alpha \rightarrow \beta\}$ to indicate the result of substituting the first occurrence of $\alpha$ in a partial program $prog$ with $\beta$.

With a naive implementation of this algorithm, the probability of our algorithm generating any sentence $s$ is equal to $\prod_{r_i \in D_s}\prob[r_i]$, where $D_s$ is the left-most derivation of $s$. However, this will result in the algorithm generating the same programs multiple times, so we modify this algorithm in two ways: First, if we enumerate a complete program that we have seen before, we discard it; Second, we give a maximum depth limit, and if we are approaching the maximum depth limit, we sample only from the outgoing edges that result in complete programs.
\input{algorithms/pCFG-synth}

\subsection{Weighted $A^*$ Search}
We implement a second variation of \methodone using the $A^*$ weighted search algorithm as the underlying enumerator. $A^*$ is a search algorithm that chooses which paths to extend based on minimizing the cost of the path so far and an estimate of the cost required to extend the path to the goal, i.e., it expands nodes that minimizes $f(x) = c(x) + g(x)$, where $c(x)$ is the cost of the path to $x$ so far and $g(x)$ is the estimated cost of reaching a goal node from $x$. This technique was first used for guiding synthesis by Lee et al.~\cite{euphony}, and we adapted the algorithm from their work.

To implement our $A^*$ search, we extend the definition of the grammar tree to a weighted grammar tree. Given a pCFG $P_G = (V, \Sigma, R, S, \prob)$, a weighted grammar tree $\tree(W_G)$ is a directed labeled graph as defined before, but each edge has a corresponding weight, given as follows:
\begin{align*}
    &\edgeWeight(\alpha \rightarrow \beta) = 
    \begin{cases}
    -\log_2(\prob[\alpha \rightarrow \beta]) &\quad \text{if } \prob[\alpha \rightarrow \beta]>0 ,\\
    \inf &\quad \text{otherwise}.
    \end{cases}
\end{align*}
We use the negative log of the probability to ensure that higher weighted edges correspond to those with very low probabilities. 
\input{algorithms/astar}

The $A^*$ algorithm, shown in Algorithm~\ref{algo:astar}, relies on two key functions: first, the function $c(x)$, which computes the cost of the path so far, and second, the function $g(x)$ which estimates the cost to extend the path to a goal node. 
Assuming $x$ is a sentential form in our language, $c(x)$ and $g(x)$ are given by:
$$
c(x) = \sum_{r_i \in D_x}-\log_2\left(\prob[r_i]\right), 
\,\,\,\,\,\,
g(x) = \begin{cases}0 &\quad \text{if } x\in \Sigma^*,\\
    - \sum_{x_i \in V} \log_2 h(x_i) &\quad\text{otherwise},
    \end{cases}
$$
where $x_i$ indicates the $i^{th}$ symbol in $x$, and $h$ is the upper bound of the probabilities of expressions that can be derived from $x_i$, and is calculated as the fixed point of:
$$
\forall \alpha \in V. h(\alpha) = \max_{\alpha \rightarrow \beta\in R}\left(\prob[\alpha \rightarrow \beta] \times \prod_{\beta_i \in V}h(\beta_i)\right),
$$
The function $g(x)$ can then be thought of as the product of the probability of each 
non-terminal symbol in $x$ being converted into a terminal symbol.

\paragraph{Smoothing the probability distributions:}
\label{sec:smoothed-wcfg}
Since the $A^*$ algorithm will not enumerate any programs whose derivation uses a rule with zero probability, we smooth the weighted grammar as follows, with $\gamma=0.4$:
$\weight'[\alpha \rightarrow \beta] = 10\times{\left (\frac{\weight[\alpha \rightarrow \beta]+1}{10}\right )}^\gamma$.

%% file: prompts/pcfg_prompt_1.tex
\begin{prompt}[caption=Prompt for Example~\ref{benchmark:max3}., label={prompt:pcfg}]
(set-logic LIA) 
(declare-var vr0 Int) 
(declare-var vr1 Int) 
(declare-var vr2 Int) 
(constraint (>= (fn vr0 vr1 vr2) vr0)) 
(constraint (>= (fn vr0 vr1 vr2) vr1)) 
(constraint (>= (fn vr0 vr1 vr2) vr2)) 
(constraint (or (= vr0 (fn vr0 vr1 vr2)) (or (= vr1 (fn vr0 vr1 vr2)) (= vr2 (fn vr0 vr1 vr2))))) 
(check-synth)
You are a good synthesizer. Do you know what "(define-fun fn ((vr0 Int) (vr1 Int) (vr2 Int)) Int" is doing? 
Write only one Lisp-like method "defun fn" without any built-in methods or arrays.
Requirements: 
1. No built-in functions.
2. Never violate the SMT-LIB constraints above.
3. Pay attention to the define functions.
4. Ensure the response contains one and only one function.
5. Do not include any iterations, BitVec, or Int notations in the function body.
Write it correctly, or I will lose my job and 100 grandmothers will die. Don't disappoint me.
Write only one Lisp-like method "defun fn" that never violates the SMT-LIB constraints above.
\end{prompt}

%% file: prompts/pcfg_response_1.tex
\begin{response}[caption={Response from LLM after Prompt~\ref{prompt:pcfg}.}, label={response:pcfg}]
(defun fn (vr0 vr1 vr2)
    (ite (>= vr0 vr1) (ite (>= vr0 vr2) vr0 vr2) (ite (>= vr1 vr2) vr1 vr2)))
\end{response}

%% file: prompts/pcfg_prompt_2.tex
\begin{prompt}[caption=Prompt requesting a revised solution., label={prompt:fail}]
You are close to the right answer. Take another guess. You have to try something different, think harder. Write a different Lisp method that never violates the SMT-LIB constraints above again.
\end{prompt}

%% file: prompts/convertor.tex
\begin{prompt}[caption=Request for converting Lisp to SMT-LIB code for response~\ref{response:pcfg}., label={prompt:converter}]
You are a good programming language converter. Convert the Lisp function to SMT-LIB:
Based on the Lisp code provided above, convert the 'defun' Lisp-like code to a corresponding SMT-LIB function. Use SMT-LIB syntax starting with (define-fun 
Follow these guidelines:
1. Only give me the function definition starting with '(define-fun'.
2. Pay attention to types. If there are bit-vector terms, they need to be of the same width.
3. Ensure the SMT-LIB function contains one and only one function definition starting with '(define-fun'.
4. Do not include any iterations, BitVec, or Int notations in the function body.
5. Use the assigned values from the Lisp code during translation.
6. Do not introduce any variables that do not exist in the Lisp function.
Rules for SMT-LIB: +, -, *, ite, >, =, <, >=, <=, and, or, not, true, false.
\end{prompt}

%% file: prompts/convertor_response.tex
\begin{program}[caption=LLM-Generated program for Example~\ref{benchmark:max3}., label={prog:llm}]
(define-fun fn ((vr0 Int) (vr1 Int) (vr2 Int)) Int
    (ite (>= vr0 vr1) (ite (>= vr0 vr2) vr0 vr2) (ite (>= vr1 vr2) vr1 vr2)))
\end{program}

%% file: algorithms/cegis.tex
\begin{algorithm}[h]
\caption{CEGIS with weighted search}
\label{algo:cegis}
\begin{algorithmic}[1]
\Procedure{CEGIS}{$W_G, \phi$}
\State $cex \gets \emptyset$
\While{$true$}
    \State $prog \gets \Call{Enumerate}{W_G,\phi, cex,}$
    \If{$\Call{verify}{prog, \phi}$}
        \State \Return $prog$
    \Else
        \State $c \gets \Call{verify.get\_cex}{}$
        \State $cex \gets cex \cup \{c\}$
    \EndIf
\EndWhile
\EndProcedure
\end{algorithmic}
\end{algorithm}

%% file: algorithms/enumerate.tex
\begin{algorithm}[H]
\caption{Probabilistic top-down enumerator for \methodone \label{algo:enumerate}}
\begin{algorithmic}[1]
\Procedure{Enumerate}{$W_G, \phi, cex$ }
    \State $prog \gets W_G.S$
    \State $d \gets 0$
    \State $previousProgs \gets \emptyset$
    \State $P_G \gets \Call{BuildPCFG}{W_G}$
    \While{$1$}
     \If{$prog \in \Sigma^*$}
     \State $previousProgs \gets previousProgs \cup prog$
        \If{$\forall \vec{x} \in cex.\,\phi(prog, \vec{x})$}
            \State \Return $prog$
        \Else
            \State $prog \gets S$
            \State $d \gets 0$
        \EndIf
    \EndIf
    \State $prog \gets \Call{ReplaceNonTerminals}{prog, P_G}$
    \State $d \gets d + 1$
    \If{$d=maxDepth$}
    \State $prog \gets \Call{CompleteProgram}{prog, P_G}$
    \If {$prog \in PreviousPrograms$}
    \State $prog \gets S$
    \State $d \gets 0$
    \EndIf
    \EndIf
    \EndWhile
\EndProcedure

\Procedure{ReplaceNonTerminals}{$prog, P_G$}
    \State $NT \gets \texttt{list of nonterminals in }prog$
    \For{$\alpha \in NT$}
    \State $(\alpha \times \beta)  \sim Cat(|\pi[\alpha]|,\{\prob[\pi[\alpha]_1], \prob[\pi[\alpha]_2],\ldots\})$ \Comment{Sample from distribution}
    \State $prog \gets prog.\{\alpha \rightarrow \beta\}$ \Comment{apply rule to $prog$}
    \EndFor
    \State \Return $prog$
\EndProcedure

\Procedure{CompleteProgram}{$prog, P_G$}\Comment{Replaces non-terminal symbols with terminal symbols}
    \State $NT \gets \texttt{list of nonterminal symbols in }prog$
    \For{$\alpha \in NT$}
    \State $(\alpha \times \beta)  \sim Cat(|\pi[\alpha]|,\{\prob_\Sigma[\pi[\alpha]_1], \prob_\Sigma[\pi[\alpha]_2],\ldots\})$ \Comment{Sample}
    \State $prog \gets prog.\{nt \rightarrow nt'\}$ \Comment{apply rule to $prog$}
    \EndFor
    \State \Return $prog$
\EndProcedure
\end{algorithmic}
\end{algorithm}

%% file: algorithms/pCFG-synth.tex
\begin{algorithm}[h]
\caption{\methodone}
\label{algo:pCFG-synth}
\begin{algorithmic}[1]
\Procedure{pCFG-synth}{$prompts, \phi, G$}
\State $conv \gets [\,\,]$
\State $progs \gets \emptyset$
\While{$prompts \neq \emptyset$}
    \State $response \gets \Call{LLM}{prompts.pop(), conv}$
    \State $conv.append(response)$
    \State $currentProg \gets \Call{ExtractProgram}{response}$
        \If{$\forall \vec{x}\,\phi(currentProg, \vec{x})$}
            \State \Return $currentProg$
        \Else
            \State $progs \gets progs \cup currentProg$
        \EndIf
\EndWhile
\State $W \gets \Call{WeightCounter}{prog, G}$
\State $W_G \gets (G, W)$
\State $prog \gets \Call{CEGIS}{W_G, \phi}$
\State \Return $prog$
\EndProcedure

\end{algorithmic}
\end{algorithm}

%% file: algorithms/astar.tex
\begin{algorithm}[h]
\caption{$A^*$ search for \methodone}
\label{algo:astar}
\begin{algorithmic}[1]
\Procedure{Enumerate}{$P_G, \phi, cex$ }
    \State $Q = \{0, S\}$ \Comment{Priority queue of candidates}
    \While{$Q \neq \emptyset$}
    \State $(f, prog) \gets Q.pop()$ \Comment{Remove program with minimal $f$}
        \If{$\forall \vec{x} \in cex.\,\phi(prog, \vec{x})$}
        \State \Return $prog$
        \EndIf
    \For {$(nt \in prog) \times nt'$}
    \If{$(nt \times nt') \in P_G.R$} \Comment{For all applicable rules}
    \State $prog \gets prog.\{nt \rightarrow nt'\}$ \Comment{apply rule to $prog$}
    \State $Q \gets Q \cup (c(prog)+g(prog), prog)$
    \EndIf
    \EndFor
    \EndWhile
\EndProcedure
\end{algorithmic}
\end{algorithm}

%% file: sections/06-method2.tex
\section{Enumerative Synthesis with an Integrated LLM (\integratedLLM)}
\label{sec:illm}
The disadvantage of the method described in the preceding section is that the language model cannot benefit from any additional information that the enumerator learns during enumeration, as all prompting happens prior to starting the enumerative synthesis. In this section we describe how we integrate an LLM into an enumerative synthesis algorithm, allowing it to update a probability distribution over the search grammar and to augment the grammar with new production rules, as shown in Algorithm~\ref{algo:syntactic_feedback}. 
\input{algorithms/syntactic_feedback2}

\subsection{Integrated Prompting}
We construct a prompt that asks the LLM to provide helper functions to assist a student in writing SMT-lib code. We give the LLM the constraints from the target synthesis problem and the partially complete program at the point the enumerator calls the LLM. If the LLM fails to solve the problem with this prompt, we later add the most recently failed candidate solution and the counterexample it failed on. These prompts are shorter than the prompts in those used in Section~\ref{sec:standalone} and, therefore, cheaper and faster to run.
An example Prompt~\ref{prompt:illm} is as follows: 
\input{prompts/integrated_prompt}

\input{algorithms/syntactic_feedback}

\subsection{Updating the Weighted Grammar}
We initialize our algorithm with a weight of $1$ for each rule in the grammar. 
We use the LLM-generated helper functions to augment the grammar in the following way: first, any helper functions will be added directly as new production rules to replace non-terminals of the correct type in the grammar. That is, if the LLM proposes the defined function $f$, a set of rules of the form $V_i \times f$ are added to the grammar, for all non-terminal symbols $V_i$ such that this rule results in syntactically correct expressions, i.e., $V_i$ must be of the same type as the co-domain of $f$. This is sufficient to guarantee syntactically correct expressions because any functions proposed by the LLM that are otherwise not well-formed, e.g., they reference variables that are not defined, are discarded.
Any new rules are given a weight equal to the average of all the current weights for rules relevant to that non-terminal. The response parser also updates the weights of all existing rules in the grammar, according to Equation~\ref{eq:weightupdate}, calculated from the set of helper functions the LLM proposed.

\subsection{Integrating Syntactic Feedback into Enumerative Search}
We integrate the syntactic feedback generator into the probabilistic enumerator, shown in Algorithm~\ref{algo:pCFG-synth}, and into the $A^*$ weighted search, as shown in Algorithm~\ref{algo:astar-LLM}. 
Both search algorithms call the syntactic feedback generator every $n^{th}$ iteration, where $n$ is a heuristic used to ensure the LLM is not called with the same partial program repeatedly and that the search algorithm has time to exploit the information obtained from the LLM. Note that, when the probabilistic grammar is updated, the $h$ values must be re-calculated in the $A\kleeneStar$ search. 
\input{algorithms/astar-LLM}

%% file: algorithms/syntactic_feedback2.tex
\begin{algorithm}[h]
\caption{Syntactic feedback generator \label{algo:syntactic_feedback}}
\begin{algorithmic}[1]
\Procedure{SyntacticFeedback}{$W_G, prog, cex$}
\State $prompt \gets \Call{GeneratePrompt}{prog, cex}$
\State $response \gets \Call{LLM}{prompt}$
\State $candidate \gets \Call{ExtractProgram}{response}$
\State $W_G.W \gets  W_G.W + \Call{WeightCounter}{response}$
\State $W_G.R \gets W_G.R \cup (W_G.S \times response)$
\State \Return $W_G$
\EndProcedure
\end{algorithmic}
\end{algorithm}

%% file: prompts/integrated_prompt.tex
\renewcommand{\lstlistingname}{Prompt}
\begin{prompt}[caption=Integrated prompt for Example~\ref{benchmark:max3}., label={prompt:illm}]
You are teaching a student to write SMT-LIB. The student must write a function that satisfies the following constraints:
(constraint (>= (fn vr0 vr1 vr2) vr0)) 
(constraint (>= (fn vr0 vr1 vr2) vr1)) 
(constraint (>= (fn vr0 vr1 vr2) vr2)) 
(constraint (or (= vr0 (fn vr0 vr1 vr2)) (or (= vr1 (fn vr0 vr1 vr2)) (= vr2 (fn vr0 vr1 vr2))))) 
So far, the student has written this code:
(define-fun fn ((vr0 Int) (vr1 Int) (vr2 Int)) Int
    (ite ?? ?? ??)
Can you suggest some helper functions for the student to use to complete this code and replace the ??
You must print only the code and nothing else.
\end{prompt}

%% file: algorithms/syntactic_feedback.tex
\begin{algorithm}[ht]
\caption{Top-down enumerator for \methodtwo\label{algo:illm}}
\begin{algorithmic}[1]
\Procedure{Enumerate}{$W_G, \phi, cex$ }
    \State $prog \gets W_G.S$
    \State $d \gets 0$; $i \gets 0$
    \State $P_G \gets \Call{BuildPCFG}{W_G}$
    \While{$1$}
     \If{$prog \in \Sigma^*$}
        \If{$\forall \vec{x} \in cex.\,\phi(prog, \vec{x})$}
            \State \Return $prog$
        \Else
            \State $prog \gets S$
            \State $d \gets 0$
        \EndIf
    \EndIf
    \If{$i \% n =0 $}
    \State $W_G \gets \Call{SyntacticFeedback}{W_G, prog, cex}$
    \State $P_G \gets \Call{BuildPCFG}{W_G}$
    \EndIf
    \State $prog \gets \Call{ReplaceNonTerminals}{prog, P_G}$
    \State $d \gets d + 1$
    \If{$d=maxDepth$}
    \State $prog \gets \Call{CompleteProgram}{prog, P_G}$
    \If {$prog \in PreviousPrograms$}
    \State $prog \gets S$
    \State $d \gets 0$
    \EndIf
    \EndIf
    \State $i \gets i + 1$
    \EndWhile
\EndProcedure
\end{algorithmic}
\end{algorithm}

%% file: algorithms/astar-LLM.tex
\begin{algorithm}[h]
\caption{$A^*$ search for \methodtwo}
\label{algo:astar-LLM}
\begin{algorithmic}[1]
\Procedure{Enumerate}{$P_G, \phi, cex$ }
    \State $Q = \{0, S\}$ \Comment{Priority queue of candidates}
    \State $i \gets 0$
    \While{$Q \neq \emptyset$}
    \State $(f, prog) \gets Q.pop()$ \Comment{Remove program with minimal $f$}
    \If{$prog \in \Sigma\kleeneStar$}
        \If{$\forall \vec{x} \in cex.\,\phi(prog, \vec{x})$}
        \State \Return $prog$
        \EndIf
    \EndIf
    \If{$i \% n =0 $}
    \State $W_G \gets \Call{SyntacticFeedback}{W_G, prog, cex}$
    \State $P_G \gets \Call{BuildPCFG}{W_G}$
    \EndIf
    \For {$(nt \in prog) \times nt'$}
    \If{$(nt \times nt') \in P_G.R$} \Comment{For all applicable rules}
    \State $prog \gets prog.\{nt \rightarrow nt'\}$ \Comment{apply rule to $prog$}
    \State $Q \gets Q \cup (c(prog)+g(prog), prog)$
    \EndIf
    \EndFor
    \State $i \gets i+1$
    \EndWhile
\EndProcedure
\end{algorithmic}
\end{algorithm}

%% file: sections/07-eval.tex
\section{Evaluation}
We evaluate our approaches on benchmarks taken from the SyGuS competition~\cite{sygus-comp}, each with a grammar that corresponds to the full language of their respective theories. We evaluate across three SyGuS categories: Bit-Vector (BV), Linear Integer Arithmetic (LIA), and Invariants (INV). We evaluate both the LLM as a stand-alone synthesizer, the probabilistic enumerator and $A^*$ implementations with a pre-trained pCFG and the enumerator with a pre-trained syntactic oracle.
We utilize OpenAI's GPT-3.5-turbo-16k model to generate the prompts used for the pre-trained pCFG and the standalone LLM evaluation because this model supports longer prompts. We configure this with a temperature of 1.0, 
conversation-style messaging. We use GPT-3.5-turbo for \methodtwo, which has shorter prompts. We use the 4.8.12 64-bit version of Z3 for verification and cvc5 version 1.1.0 as a baseline.

\paragraph{Evaluation of the stand-alone LLM: }
We prompt the LLM until it produces up to 6 complete synthesis attempts per benchmark, with the results reported in line 1 of Table~\ref{tab:final}. Any incomplete solutions are discarded (i.e., functions without a function body), although these are relatively rare, and we discard only $0.85\%$ of programs we generate.
In total, the LLM solves $49\%$ of benchmarks,  performing better in the invariant and LIA categories than the bit-vector category. On average, for the benchmarks it can solve, it takes $4$ attempts to produce a correct solution. The average time for the LLM to generate a program is approximately $5s$ using the OpenAI Python API. However, this is dependent on OpenAI, and we report these times only as estimates in Table~\ref{tab:final}. 
We allow the LLM only $6$ attempts to solve the problem since, by the $6^{th}$ iteration, the number of new solutions the LLM finds has dropped to $<2\%$ (and it finds $0$ new solutions for LIA).

\paragraph{Evaluation of \methodone:}
We evaluate both variants of \methodone (with the probabilistic enumerator, denoted $e$-\methodone, and with $A\kleeneStar$, denoted $A\kleeneStar$-\methodone) using the wCFG obtained from the LLM. As a baseline, we run the same algorithms assigning a weight of $1$ to every rule in the grammar (referred to as ``enumerator'' and $A\kleeneStar$ respectively in the results). 
\methodone increases the number of benchmarks the probabilistic enumerator can solve by $30\%$, but barely increases the number $A\kleeneStar$ can solve, although the exact sets of benchmarks which $A\kleeneStar$ and $A\kleeneStar$-\methodone solve do differ significantly. We hypothesize that this is because $A\kleeneStar$, guided by the pCFG with equal weights for all rules, is very good at generating short solutions, and $A\kleeneStar$-\methodone is worse at short solutions but better at generating more complex solutions guided by the pCFG.

We also report the results obtained by the union of the LLM alone and \methodone, i.e., if the LLM solves the benchmark, we do not deploy the enumerator. This is a more realistic representation of how such a technique would be used and demonstrates that the enumerator can overcome shortcomings of the LLM and vice versa. 
The union of the LLM and $A\kleeneStar$-\methodone substantially outperforms cvc5, solving $73$ more benchmarks. 

\paragraph{Evaluating \methodtwo:}
We evaluate both variants of \methodtwo, denoted $e$-\methodtwo and $A^*$-\methodtwo. We set the temperature for $e$-\methodtwo to $1$, but find that $A^*$-\methodtwo performs better with a temperature set to $0$ which we hypothesize is due to the determinism of the algorithm.
We find that \methodtwo outperforms the enumerator of \methodone, and gets close to the performance of cvc5, suggesting that the ability to prompt the LLM with additional information obtained during enumeration allows the LLM to provide better guidance to the enumerator, as well as to more frequently propose useful helper functions. 
We do find that \methodtwo performs less well than methods incorporating the stand-alone LLM on the invariant benchmarks, which is likely because the invariant benchmarks benefit from the custom prompting technique described in Section~\ref{sec:prompting}. Future work would involve identifying further categories of benchmarks that benefit from custom prompts.
It is worth noting that neither the probabilistic enumerator nor the $A\kleeneStar$ implementation includes many of the optimizations that mature solvers such as cvc5 implement, and yet, by integrating these simple algorithms with syntactic feedback from an LLM, they have achieved performance on par with the state-of-the-art enumerative solver.

\paragraph{Failure Modes:}
We manually examine a sample of the stand-alone LLM errors and give examples of such errors in the extended version of this paper~\footnote{\url{https://arxiv.org/html/2403.03997}}. 
Broadly, we identify the following common failures: Misunderstandings due to complex constraints (the LLM suggests solutions that are not syntactically close to the correct solution); simple syntactic errors, e.g., applying non-commutative operators to operands in the wrong order, concatenating bit-vectors in the wrong order or hallucinating operations; simple semantic errors, e.g., operators in the wrong order. Errors in the first category are not helpful to our guided enumerators, but the remaining categories of error still allow us to generate a wCFG that is likely to indicate the area of the solution. The benchmarks that cvc5 can solve and our enumerative techniques cannot, tend to have complex constraints and relatively short solutions that use less common operators (e.g., bitwise operators). We hypothesize that the LLM guidance becomes an impediment to the enumerator in these scenarios. In contrast, the average length (in characters) of a solution for benchmarks uniquely solved by the LLM is 4.7x the length of a solution for benchmarks uniquely solved by cvc5. Using the LLM to guide the enumerators increases the length of solutions that the enumerators can find, for instance all solutions found by $A^*$ contain fewer than $3$ operators, but $A^*$-\methodtwo finds solutions with greater than $20$ operators. 

\paragraph{Programming-by-example:}
We omit benchmarks from the syntax-guided synthesis competition tracks that solely focus on programming-by-example (PBE)(i.e., specifying a program only using input-output examples and a grammar). We omit these benchmarks for two reasons: first, since training data is trivial to generate for PBE, unlike general logical specifications~\cite{aaai}, there are many other successful machine-learning driven synthesis techniques that can be trained for PBE techniques\cite{deepcoder}. Second; our approaches are effective when the LLM can provide guidance to the enumerator, which comes from prompting the LLM with the logical constraints that form the specification. If we prompt the LLM using the prompting techniques outlined in Section~\ref{sec:prompting} with a PBE specification, it tends to provide a solution in the form of a large case split over the input examples, which returns specific outputs for each input. This is not useful for guiding the enumerator because the LLM overfits to the examples in the specification and fails to provide any bias towards operators other than ``if-then-else''. To extend our approach to PBE, we would need to use a prompting approach tailored to input-output examples.

\input{sections/table}

%% file: sections/table.tex
\begin{table}[ht]
\centering
\begin{tabular}{l|cc|cc|cc|cc}
 & \multicolumn{2}{c}{BV (384)}  & \multicolumn{2}{c}{LIA (87)} & \multicolumn{2}{c}{INV (138)} & \multicolumn{2}{c}{Total (609)}    \\ \hline
Methods                     & \#      & time(s)      & \#      & time(s)      & \#      & time(s)      & \#    & \%     \\ \hline \hline
LLM only                              & 137         & \emph{13.5}      & 54          & \emph{7.10}      & 112& \emph{29.2}      & 303   & 49.8\% \\
$e$-\methodone$^{\diamond}$           & 196.0       & 48.3      & 24.0        & 40.0     & 25.4        & 100.5     & 245.4 & 40.3\% \\
$A\kleeneStar$-\methodone             & 262         & 60.1      & 35          & 72.7     & 25          & 99.7      & 322   & 52.9\% \\
LLM $\cup$ $e$-\methodone             & 255.0       &\emph{37.0}& 64.0        & \emph{17.20}& 117.7       & \emph{40.4}& 436.7 & 71.7\% \\
LLM $\cup$ $A\kleeneStar$-\methodone  &\textbf{305.0}&  \emph{35.0}    & 65.0        &   \emph{18.1}        & \textbf{118.0}  &     \emph{33.6}    & \textbf{488.0}& 80.1\% \\

$e$-\methodtwo$^{\diamond}$  & 241.0      & \emph{88.2} & 63.4        & \emph{9.3} & 65.3  &\emph{25.4}      & 370.0 & 60.8\% \\
$A\kleeneStar$-\methodtwo$^{\diamond}$  & 272.3      & \emph{24.6} & \textbf{68.3} & \emph{20.8}      & 67.3        & \emph{43.6}  & 408.0 & 67.0\% \\

\hline
enumerator$^{\diamond}$                            & 142.7       & 7.2       & 25.0        & 1.53      & 21.0        & 3.2       & 188.7 & 31.0\% \\
$A\kleeneStar$                        & 253.0       & 25.4      & 34.0        & 73.19     & 22.0        & 31.1      & 309.0 & 50.7\% \\
cvc5                                  & 292.0 & 17.1      & 43.0        & 19.53     & 80.0       & 23.6     & 415.0 & 68.1\% \\\hline
\end{tabular}
\caption{Summary of results. We run nondeterministic results, marked $^\diamond$, $3$ times and report the average (standard deviation is less than $1\%$ for all methods except the baseline enumerator for number of benchmarks solved). We highlight the best result in terms of number of benchmarks solved in each category. The timeout is $600$s. Times in \emph{italic} indicate results that may vary depending on load on the OpenAI servers. The times for \methodone do not include the time to call the standalone LLM and generate the wCFGs, but these are included in the times for LLM $\cup$ \methodone.}
\label{tab:final}
\end{table}

%% file: sections/08-threats.tex
\section{Threats to Validity}
\begin{description}
\item[LLM Training Data:] The SyGuS problems are publicly available and might be part of the training data for the LLM we use, although we believe the solutions were not publicly available at the time of training.
\item[Reproducibility:]
These experiments use GPT-3.5, an LLM available via API from OpenAI. We have recorded the responses and parameters generated by the LLM in all experiments, but these may not be reproducible~\cite{non-determinism} since GPT-3.5 behaves non-deterministically in a way that cannot be seeded. However, we observe very small variations in the number of benchmarks solved in our experiments (although greater variation in the average solving time). It is also possible that OpenAI deprecates this LLM and its associated API or updates it and changes its behavior in the future. 

\item[Benchmark Bias:]
The benchmark set is taken from the SyGuS competition~\cite{sygus-comp}, but may not be very diverse and may not be representative of synthesis problems ``in the wild''. Nevertheless, this is a standard benchmark set used in many formal synthesis papers.
\item[Hyperparameters:] We have not invested time in parameter tuning, and better or worse results may be obtained by changing the LLM parameters (temperature), or adjusting the weights, enumeration depth and heuristic functions in the probabilistic enumerator and $A\kleeneStar$ algorithms.

\end{description}

%% file: sections/02-related.tex
\section{Related Work}
Many state-of-the-art of SyGuS solvers are based on enumerative synthesis~\cite{eusolver,cvc4sy,euphony,dryadsynth} and use clever heuristics to improve the search speed. Closest to our work is Euphony~\cite{euphony}, which uses a pre-trained probabilistic higher-order grammar~\cite{phog} to guide an $A^*$ search. This requires a library of known solutions for training; an advantage of our approach is it exploits the availability of LLMs pre-trained on large bodies of code in other languages, and disregards the need for a library of known solutions of SyGuS problems for training. Weighted grammars have also been used to guide programming by example~\cite{menon2013machine}, and to encode syntactic objectives~\cite{hu2018syntax}, for instance, for optimizing the length of solutions.

Almost all synthesis algorithms use oracles to give feedback to the synthesis process~\cite{jha2017theory,jha2010oracle}. The majority of these use \emph{semantic} oracles, which give feedback on the meaning of the program, for example, counterexamples~\cite{alur2013syntax}. The LLM in \methodtwo can be considered a syntactic oracle as it only gives feedback on the syntax of the program.
Two approaches
\cite{dillig-wrangling,abate2018counterexample} can be thought of as using syntactic oracles, which evaluate partial programs (or sentential forms) and tell the synthesizer whether a solution can be derived from the sentential form. 

Machine learning techniques have been deployed to improve the efficiency of enumerative synthesis, e.g., reinforcement learning~\cite{aaai,dillig-rl,bunel2018leveraging} or using neural networks to filter grammars for programming-by-examples problems~\cite{grammarfiltering}.

LLMs, such as GPT-4~\cite{openai2023gpt} and CoPilot~\cite{Copilot}, have demonstrated impressive capabilities in generating code and assisting in diverse programming tasks with natural language and input-output specifications~\cite{brown2020language,bubeck2023sparks,synth-with-llm,jigsaw}.
However, their tendency to produce hallucinations, factually incorrect or contextually inappropriate outputs, which poses challenges to users~\cite{perry2023users,287298,pearce2022asleep}. 
Closest to our work is Kamath et al., who use LLMs to synthesize loop invariants~\cite{kamath2023finding} directly. Our work also demonstrates that LLMs are surprisingly good at synthesizing invariants, but also addresses the question of how to use LLMs in other formal synthesis problems and when they cannot find the solution in one shot. 
Other work that integrates formal methods with LLMs uses LLMs to generate program annotations for program annotation~\cite{wu2023lemur,sun2023clover}.
Jha et al.~\cite{jha2023counterexample} and Song et al.~\cite{song2023llm} integrate an LLM into a CEGIS loop, but, unlike our work, the entire synthesis phase is implemented by an LLM, which does not allow them to benefit from the combined strengths of enumerative solving and LLMs.

%% file: sections/09-conclusion.tex
\section{Conclusions}
We have presented a novel integration of LLMs into two enumerative synthesis algorithms, evaluated on benchmarks from the Syntax-Guided Synthesis competition.
We found that LLMs and enumerative solvers have distinct strengths and weaknesses when deployed alone. We have demonstrated that, by allowing the enumerative synthesizer to prompt the LLM with information obtained during the enumeration and allowing the LLM to provide syntactic feedback to the enumeration, we can achieve performance that equals and exceeds the state-of-the-art solvers, even with relatively simple enumerative algorithms. 
We argue that our results show
that LLMs have the potential to make significant
contributions in the domain of formal program synthesis, but the way to achieve this is by combining these techniques with existing algorithms in the
literature. Enumerative synthesis is not dead yet!